\documentclass[conference]{IEEEtran}
\IEEEoverridecommandlockouts
\usepackage{cite}
\usepackage{amsmath,amssymb,amsfonts}
\usepackage{algorithmic}
\usepackage{graphicx}
\usepackage{textcomp}
\usepackage{xcolor}
\def\BibTeX{{\rm B\kern-.05em{\sc i\kern-.025em b}\kern-.08em
    T\kern-.1667em\lower.7ex\hbox{E}\kern-.125emX}}
    
\begin{document}

\title{Defense-PointNet: Protecting PointNet Against Adversarial Attacks
}

\author{\textbf{Yu~Zhang\textsuperscript{1}*,
		Gongbo~Liang\textsuperscript{1},
		Tawfiq~Salem\textsuperscript{2},
		Nathan~Jacobs\textsuperscript{1}
		 } \\ [1ex]
		$1$ Department of Computer Science, University of Kentucky, Lexington, KY, USA\\
		$2$ Department of Computer Science and Software Engineering, Miami University, Oxford, Ohio, USA \\
		Email: y.zhang@uky.edu*}

\maketitle

\begin{abstract}
    Despite remarkable performance across a broad range of tasks, neural networks have been shown to be vulnerable to adversarial attacks.
    Many works focus on adversarial attacks and defenses on 2D images, but few focus on 3D point clouds. 
    In this paper, our goal is to enhance the adversarial robustness of PointNet, which is one of the most widely used models for 3D point clouds.
    We apply the fast gradient sign attack method (FGSM) on 3D point clouds and find that FGSM can be used to generate not only adversarial images but also adversarial point clouds. To minimize the vulnerability of PointNet to adversarial attacks, we propose Defense-PointNet.
     We compare our model with two baseline approaches and show that Defense-PointNet significantly improves the robustness of the network against adversarial samples.
\end{abstract}


\section{Introduction}
Convolutional neural networks (CNNs) achieve remarkable performance across a broad range of image-related tasks~\cite{pmlr-v97-tan19a}~\cite{law2018cornernet}~\cite{ronneberger2015u}, but CNNs have been shown to be vulnerable to adversarial attacks~\cite{su2019one}~\cite{moosavi2016deepfool}~\cite{carlini2017towards}~\cite{papernot2016distillation}. Goodfellow \textit{et al.}~\cite{goodfellow2014explaining} propose the fast gradient sign attack method (FGSM) for generating adversarial samples and claim that CNNs can be easily misled with high confidence by adding imperceptible perturbations to real input images. Recently, researchers propose multiple ways to generate adversarial samples for 2D images and explain this phenomenon theoretically~\cite{szegedy2013intriguing}~\cite{ papernot2016limitations}. Besides generating adversarial samples, some recent works focus on how to protect CNNs against adversarial attacks. ShieldNets~\cite{theagarajan2019shieldnets} proposes a probabilistic method to defend against adversarial attacks on 2D images. Defense-GAN~\cite{samangouei2018defense} models the distribution of unperturbed images, and Erraqabi \textit{et al.}~\cite{erraqabi2018a3t} propose an adversarially augmented adversarial training (A3T) approach to improve the adversarial robustness of CNNs by using a discriminator to filter the adversarial noise and achieve good performance on MNIST~\cite{deng2012mnist}.

\begin{figure}
  \centering
  \includegraphics[width=0.5\textwidth]{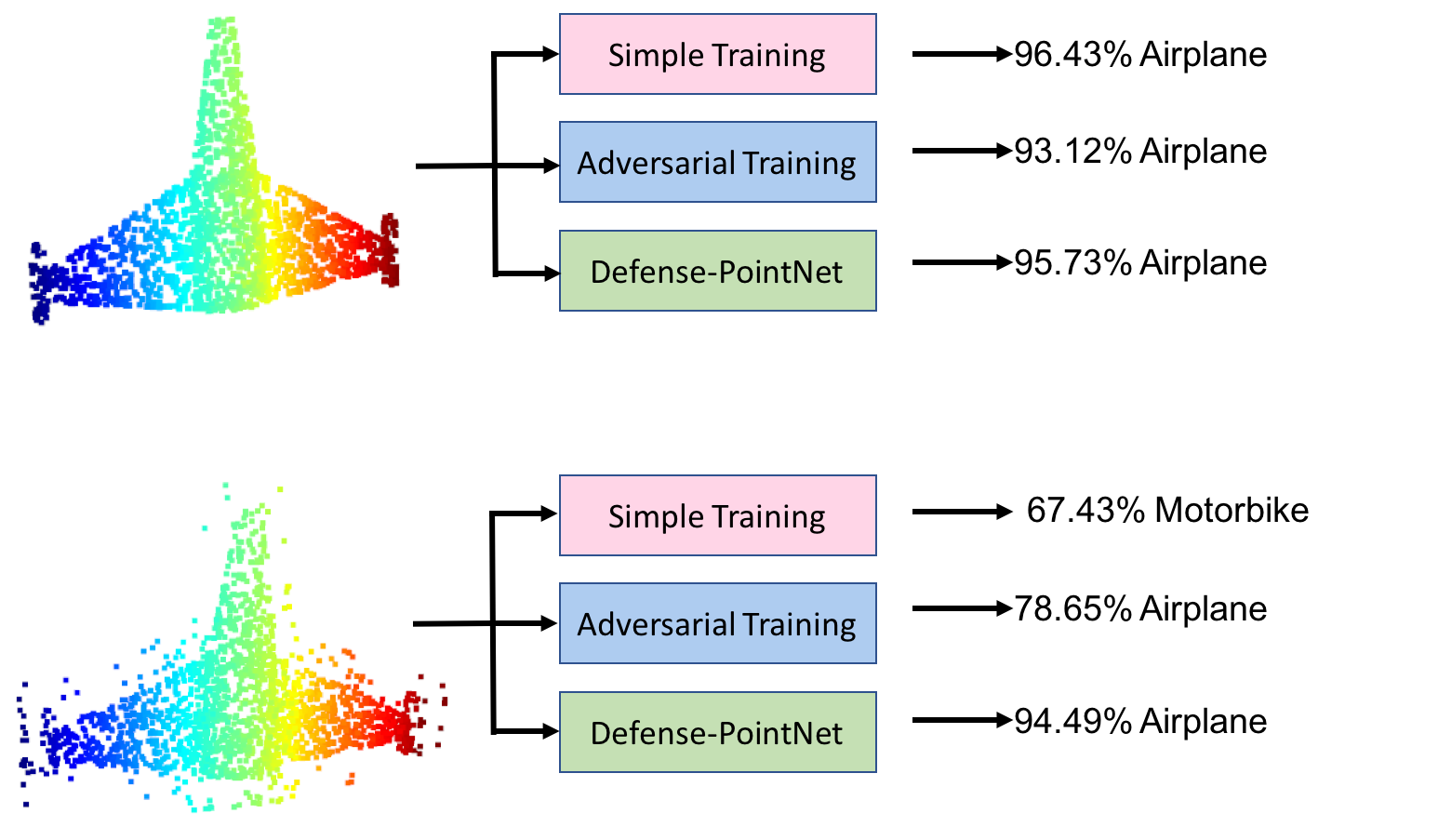}
  \caption{The upper sample is a clean airplane point cloud and the lower sample is its adversarial counterpart generated by FGSM. Both samples are fed into three models. For the real sample, all three approaches classify correctly but for the adversarial sample, it fools the simple training model.}
  \label{figs:cartoon}
\end{figure}
Many works focus on adversarial attacks and defenses on 2D images, so one question naturally arises: How can we transfer these attacking and defending approaches from 2D images to 3D real world data? There are many ways to represent 3D objects in the real world. In this work, we focus on 3D point clouds, which can be obtained from LiDARs and depth cameras. Some recent works~\cite{qi2017pointnet}~\cite{qi2017pointnet++}~\cite{wang2018dynamic} use neural networks to process point cloud data and achieve great success. Among these works, PointNet~\cite{qi2017pointnet} is the earliest and also the most commonly used architecture for 3D point cloud data, and we show that it can be attacked by adversarial samples. Figure~\ref{figs:cartoon} illustrates that when the PointNet is mislead by the adversarial sample, our model can still make the correct prediction.

Previous works~\cite{qi2017pointnet}~\cite{xiang2019generating} show that PointNet is more difficult to attack than ordinary CNNs. 
It has been proven that PointNet is robust against multiple kinds of perturbations. We validate this by showing that PointNet can achieve high accuracy as 86.57\% when attacked by FGSM with $\epsilon=0.1$ and 64.78\% with $\epsilon=0.2$. One major difficulty for generating adversarial point clouds is that we cannot simply use most of the attacking approaches designed for 2D images because it is impossible for us to modify pixel values for point clouds unless we have point-level features. Xiang \textit{et al.}~\cite{xiang2019generating} propose multiple approaches to generate adversarial point clouds by shifting points or adding extra points to the original point cloud. In this work, we focus more on how to improve the adversarial robustness than how to generate adversarial samples.

To protect PointNet against adversarial attacks, we propose Defense-PointNet which uses a discriminator to learn to filter the adversarial noise in the latent representation space. We split the PointNet into two parts, the feature extractor and the classifier. We attach the discriminator to the last layer of the feature extractor and train these three modules jointly. Our discriminator is trained to distinguish latent representations of real point clouds from the ones of adversarial point clouds. The feature extractor is trained not only to extract features for the classifier to correctly classify the training samples but also to fool the discriminator. 

 We evaluate our Defense-PointNet on ShapeNet~\cite{chang2015shapenet}, a well-known 3D point cloud dataset, and find that it outperforms two baseline approaches and improves the adversarial robustness of PointNet. We also find that Defense-PointNet can give higher probability on correctly classified samples comparing with the traditional adversarial training approach, which means Defense-PointNet can predict not only more accurately but also more confidently. We apply t-Distributed Stochastic Neighbor Embedding (t-SNE)~\cite{maaten2008visualizing} for dimensionality reduction and use it to visualize the latent vectors. Our experiments show that Defense-PointNet can enhance the intra-class compactness of feature clusters, thereby reducing the overlap of different classes, leading to the improvement of robustness against bounded input perturbations. The key contributions of this work are:
\begin{itemize}
  \item Adversarial Point Clouds Generation: We apply fast gradient sign attack method to point clouds and find that even slightly shifting the points can mislead PointNet to give incorrect predictions with high confidence.
  \item Adversarial Training: By feeding real batches and adversarial batches alternatively in the network during training, the adversarial robustness of PointNet is improved significantly. We use this adversarial training approach as one baseline for evaluation.
  \item Defense-PointNet: We propose the Defense-PointNet which can protect PointNet against adversarial attacks. By adding a discriminator, we enforce resistance to adversarial attack of latent representations and improve the adversarial robustness of PointNet.
  \item Visualization: We provide t-SNE plots to explain how our approach affects the feature space representations.
\end{itemize}


\begin{figure*}
  \centering
  \includegraphics[width=0.9\textwidth]{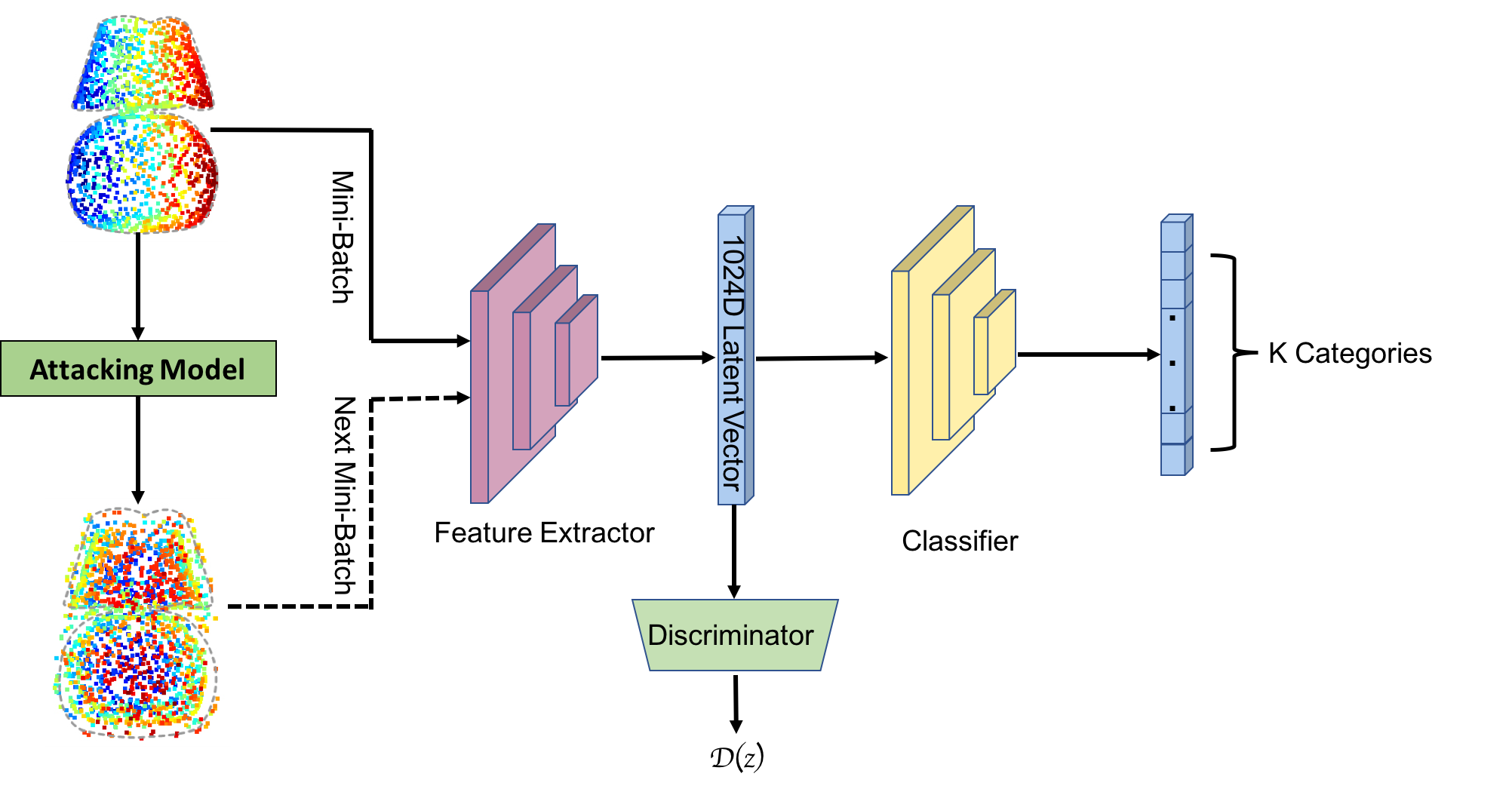}
  \caption{The Architecture of Defense-PointNet}
  \label{figs:net}
\end{figure*}

\section{Related Work}

\subsection{Fast Gradient Sign Attack Method}
Fast gradient sign attack method (FGSM)~\cite{goodfellow2014explaining} is the earliest and most fundamental technique for generating adversarial samples. 
Many following attacking methods, for instance, Basic Iterative Method (BIM)~\cite{kurakin2016adversarial}, Momentum Iterative Method (MIM)~\cite{dong2018boosting}, Carlini \& Wagner Attack (C\&W)~\cite{carlini2017towards}, and Projected Gradient Descent (PGD)~\cite{madry2017towards} are all based on FGSM or using FGSM as one of their steps. In FGSM, we calculate the gradient of the cost with respect to the input pixels and generate a perturbation matrix with same dimension of the input images. By adding the perturbation matrix to the input images, we can generate adversarial images with limited perturbations. The adversarial samples it generates can mislead the CNNs to predict incorrect labels with high confidence.  Our main focus is to provide a defending strategy against the fundamental but powerful FGSM. In our paper, we directly apply FGSM on 3D point clouds, which means we shift points instead of modifying pixel values.

\subsection{Defense-GAN}
Defense-GAN~\cite{samangouei2018defense} is a new framework leveraging the expressive capability of generative models to defend deep neural networks against adversarial attacks. Defense-GAN is trained to model the distribution of unperturbed images. At inference time, it finds a close output to a
given image which does not contain the adversarial changes, then feed this image to the classifier. It has been proven that Defense-GAN is effective against different attack approaches and improves on existing defending strategies.

\subsection{A3T}
The original idea of augmenting a network with a discriminator to make hidden representation filter the noise is the essence of the work Ganin \textit{et al.}~\cite{ganin2016domain} on domain adaptation, where the network learns features
that adapt to different domains for the same task. That idea is exploited by Adversarially Augmented Adversarial Training (A3T)~\cite{erraqabi2018a3t}, which divides a CNN into an encoder and a residual classifier. A discriminator is used to enforce resistance to adversarial attack of latent representations. The A3T approach has been evaluated on MNIST and achieved higher accuracy than using the adversarial training approach. The success of these two works~\cite{ganin2016domain, erraqabi2018a3t} on 2D images inspire us to augment deep neural networks (DNNs) for 3D point cloud data and design the Defense-PointNet.

\subsection{PointNet}
Unordered point sets in 3D are usually considered to be difficult to model by using DNNs. PointNet~\cite{qi2017pointnet} is the first paper using DNNs to model 3D point cloud data. PointNet uses max pooling~\cite{cirecsan2012multi} to reduce the unordered and varying length input to a fixed-length global feature vector make it possible for end-to-end training. In our paper, we show that PointNet can still be attacked using adversarial samples generated by the fundamental approach FGSM, and we propose the Defense-PointNet to protect PointNet against adversarial attacks.

\subsection{t-SNE}
The t-Distributed Stochastic Neighbor Embedding (t-SNE)~\cite{maaten2008visualizing} that we used is a technique for dimensionality reduction that is particularly well suited for the visualization of high-dimensional datasets. The technique can be implemented via Barnes-Hut approximations, allowing it to be applied on large real-world datasets. t-SNE has been approved to be able to apply on data sets with up to 30 million examples~\cite{van2014accelerating}~\cite{van2012visualizing}~\cite{van2009learning}. In this work, we use t-SNE to map our high dimensional latent vectors to 2D space for visualization. That helps us explain how our model affects the feature space representations.

\section{Approach}
We first use FGSM to attack the PointNet. Then we propose the architecture of our Defense-PointNet. We optimize the parameters of our model by minimizing three loss functions: the classifier loss, the discriminator loss, and the feature extractor loss simultaneously. 

\subsection{Adversarial Point Clouds Generation}
We generate adversarial point clouds by shifting points using FGSM. 

Our PointNet takes as input a mini-batch of real point clouds $x$ and its associated targets $y$. We then calculate the gradients of that batch of point clouds and add the gradients to $x$ to get perturbed point clouds as adversarial samples $x_{p}$ as following:
\begin{equation}
    x_{p} = x + \epsilon \nabla_{x}(\theta, x, y). \nonumber
\end{equation}
By adding a small perturbation to the input point clouds $x$, we get a new mini-batch of point clouds $x_{p}$, which is shifted slightly from $x$ and we use $x_{p}$ as our adversarial point clouds.

\subsection{Adversarial Training}
In this part, we train our model on a mixture of clean and adversarial samples. Specifically, for each iteration, we first feed a mini-batch of real point clouds to the network, then generate and feed the corresponding mini-batch of adversarial point clouds alternatively.

\subsection{Denfense-PointNet}
We extend the adversarial training procedure by proposing the Defense-PointNet. We split the PointNet into two parts. The first part is the feature extractor and the second part is the classifier. A discriminator is attached to the last layer of the feature extractor. Figure~\ref{figs:net} illustrates the architecture of the Defense-PointNet.

The classifier is trained to classify each input correctly and the feature extractor is trained to not only extract features for the classifier but also fool the discriminator. The output of the feature extractor is a $1024D$ latent vector. That latent vector is the input of the discriminator, which is a two-layer network to enforce an invariance across real samples and their adversarial counterparts at the level of the latent representations. If a latent vector is extracted from a real point cloud, it is labeled as $t=0$, and if it is extracted from an adversarial sample, then it is labeled as $t=1$. The discriminator is trained as a binary classifier and it learns to output the probability of the input latent vector's tag is $t=0$ .

We then use the response of the discriminator to train the feature extractor. The feature extractor is trained to fool the discriminator and try to mislead the discriminator to label every real/adversarial vector as real ($t=0$). 

\subsection{Loss Function}
To optimize the parameters of Defense-PointNet, we design three different loss functions respectively for the classifier, the discriminator, and the feature extractor. We optimize these three loss functions simultaneously during training. 

\bigbreak

\noindent\textbf{Classifier Loss:}
Our first loss, $L_{cls}$, is the loss of the classifier. The purpose of the classifier is for multi-class classification. We use negative log likelihood as our loss function here as the following equation:
\begin{equation}
 L_{cls}(x_{pred}, y) = -\log P(y|x_{pred}). \nonumber
\end{equation}
The inputs are the prediction vector $x_{pred}$ and the target $y$. Notice that $L_{cls}$ is used to update parameters in the entire PointNet, including both the classifier and the feature extractor.
\bigbreak
\noindent\textbf{Discriminator Loss:}
Our second loss, $L_{dis}$, is the loss of the discriminator. For every latent vector $z$, the discriminator predicts the probability $D(z) = P(t=0|z)$ that $z$ is from real ($t=0$). We use binary cross entropy for the discriminator loss as following: 
\begin{equation}
\begin{split}
L_{dis}(D(z), t) = & -\dfrac{1}{n}\sum_{i=1}^{n}[t^{i}\log({D(z)^{i}})+ \\
&(1-t^{i})\log(1-{D(z)^{i}})]. \nonumber
\end{split}
\end{equation}
For the real mini-batch, we use $t=0$; for the adversarial mini-batch, we use $t=1$.
\bigbreak
\noindent\textbf{Feature Extractor Loss:}
The goal of our feature extractor is not only to extract features for the classifier but also to fool the discriminator. We use the response from the discriminator to train the feature extractor. We use binary cross entropy loss here for the feature extractor as same as what we use for $L_{dis}$, the only difference is for $L_{feat}$, we always use $t=0$ for both real and adversarial mini-batches. The loss function is shown in the following equation:
\begin{equation}
\begin{split}
L_{feat}(D(z), t=0) = & -\dfrac{1}{n}\sum_{i=1}^{n}[t^{i}\log({D(z)^{i}})+\\
&(1-t^{i})\log(1-{D(z)^{i}})]. \nonumber
\end{split}
\end{equation}
$L_{feat}$ is only used to update parameters in the feature extractor.

\section{Evaluation}
We evaluate our proposed approach quantitatively and qualitatively. We begin by introducing the dataset we use, then we show the implementation details. Finally we show the results of our model and compare our results with different baseline approaches.

\begin{figure} 
  \centering
  \includegraphics[width=0.45\textwidth]{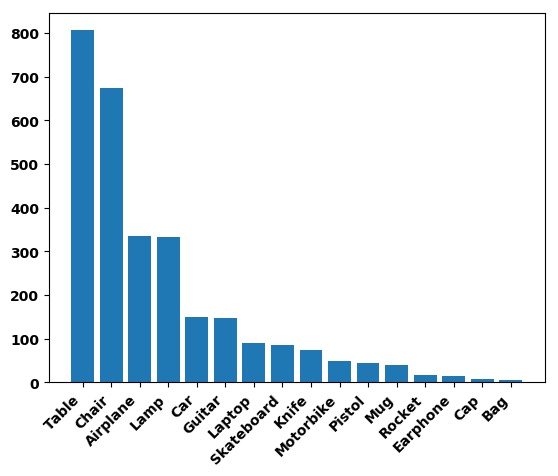}
  \caption{Training Set Distribution}
  \label{figs:training set}
\end{figure}
\subsection{Dataset}
\label{sec:dataset}
To evaluate our approaches, we use ShapeNet dataset~\cite{chang2015shapenet}, which is one of the most widely used benchmark datasets of 3D point clouds. It is a richly-annotated, large-scale repository of shapes represented by 3D CAD models of objects. In our evaluation, we use a subset, which contains 15,011 3D point clouds belongings to 16 categories. We split the dataset to 80\% training and 20\% testing. This results in 12,137 samples for training and 2,874 samples for evaluating. The training set distribution is showed in Figure~\ref{figs:training set}. 

\begin{figure*}
  \centering
  \includegraphics[width=1.0\textwidth]{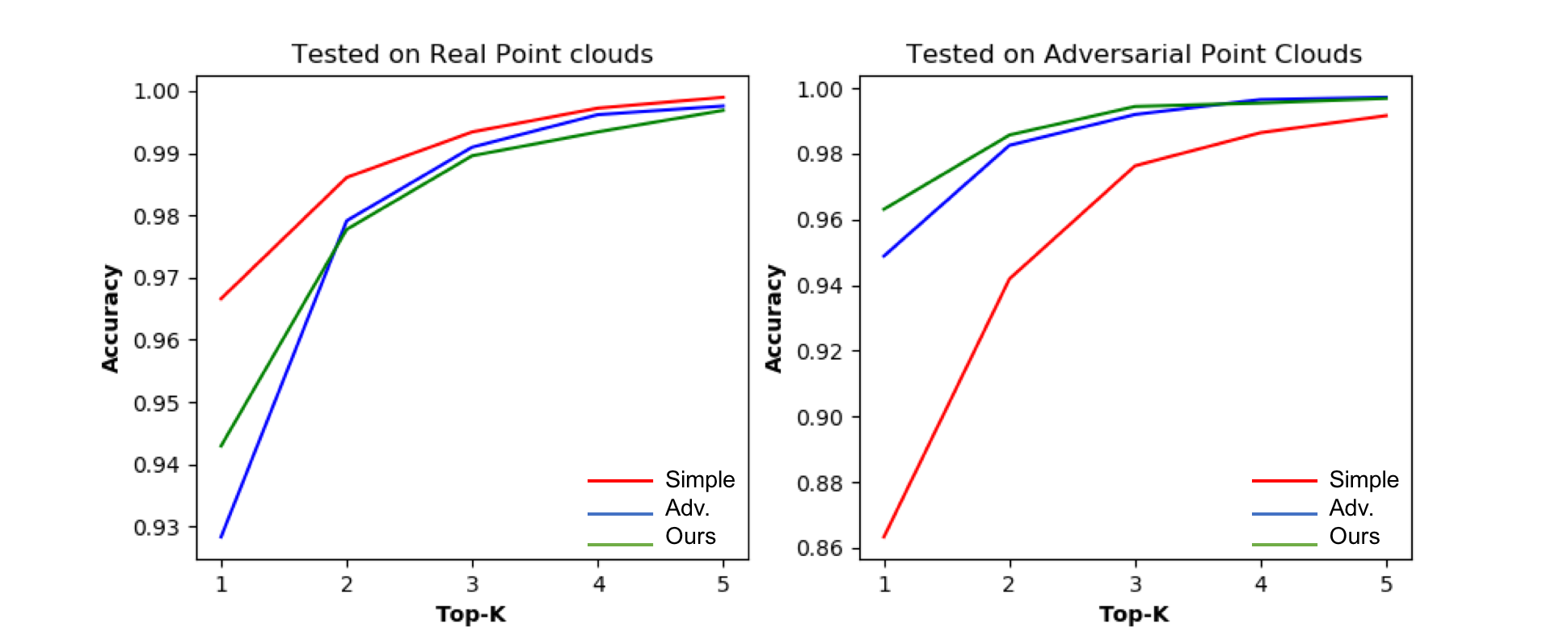}
  \caption{Top-K accuracy: In each subfigure, the horizontal axis stands for top-K (K in the range $[1, 5]$) and the vertical axis stands for top-K accuracy, which means that the correct prediction gets to be in the top-K probabilities for it to count as correct. (left) Shows all three approaches have similar performance when testing on real samples. (right) Shows our approach outperforms the other two baseline approaches when testing on adversarial samples.}
  \label{figs:top5}
\end{figure*}

\begin{figure*}
  \centering
  \includegraphics[width=1.0\textwidth]{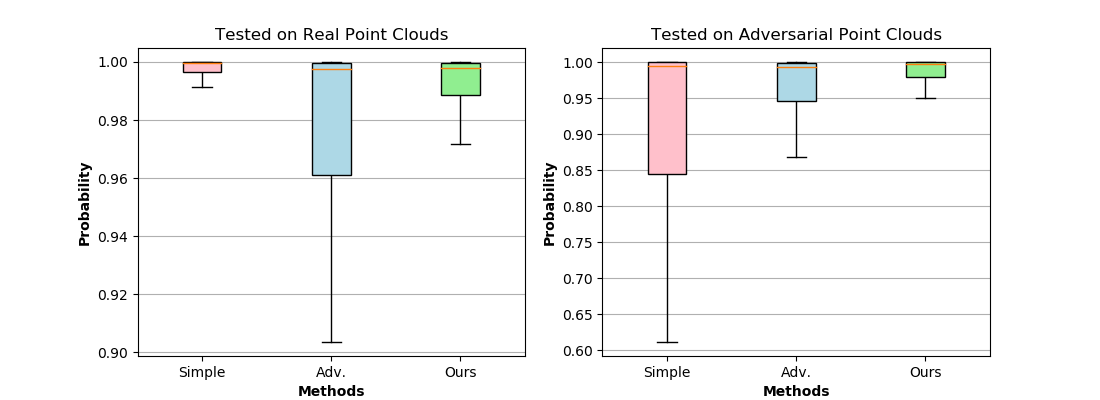}
  \caption{Two box plots, which show the prediction confidence of the proposed approach and the two baseline approach on the real (left) and adversarial (right) samples.}
  \label{figs:boxplot}
\end{figure*}

\subsection{Implementation Details}
Our approach was implemented in Pytorch~\cite{paszke2017automatic} and trained and tested on a Linux computer server with two Nvidia GTX $1080$ GPUs. For the FGSM, we use $\epsilon = 0.1$. We split the PointNet into a feature extractor and a classifier. The classifier is a combination of the last three fully connected layers of the PointNet and we set the output dimension of FC3 to 16, which is the number of the classes. Anything before the classifier in the PointNet architecture is considered as the feature extractor. The output of the feature extractor is a $1024D$ latent vector. The discriminator is a combination of 2 linear layers which is attached to the end of the feature extractor. The dimension of the discriminator's input is 1024 and the output dimension is 2. We use softmax as the activation function to predict the probability of the latent vector is real ($t=0$). We use Adam~\cite{kingma2014adam} as our optimizer and decay the learning rate of each parameter group by $~\gamma=0.5$ every 20 epochs.


\begin{table}[!b]
\caption{Accuracy on Test Set}
\centering
\begin{tabular}{@{}cccll@{}}
\hline
Methods              & Acc. on real     & Acc. on adversarial   \\ 
\hline
Simple Training      & 96.62\%          & 86.57\%               \\
Adversarial Training & 93.04\%          & 94.92\%               \\
Defense-PointNet   & 94.08\%          & \textbf{96.35\%}      \\
\hline
\end{tabular}
\label{tab:results}
\end{table}

\subsection{Classification Accuracy}
We use simple training and adversarial training approaches as our two baselines. Simple training means we train the PointNet model only on real data (the original ShapeNet data). For the adversarial training, we train the model on the mixture of the original ShapeNet data and the adversarial data generated by us used FGSM. We compare the evaluation results of our approach with these two baselines, and our experiments show the proposed Defense-PointNet outperforms both of these two baselines on testing accuracy. 

Table~\ref{tab:results} shows the quantitative testing results on the three compared models. The table reveals when evaluating these approaches on the real samples, all of them achieve reasonably good performance (the accuracy between $93.04\%$ and $96.62\%$). However, when testing on the adversarial samples, the accuracy of the simple training approach decreases significantly (from $96.96\%$ to $86.57\%$).  The adversarial training slightly increases its accuracy to $94.92\%$, and the performance of our approach is significantly improved to $96.35\%$. Figure~\ref{figs:top5} shows the top-1 to top-5 accuracy for all the three approaches on both real and adversarial testing samples. The figure also indicates our approach outperforms the other two baseline approach when testing on adversarial samples.

Figure~\ref{figs:top5} (left) shows that our approach does not outperform the simple training approach, but this results are based only on real samples, not on adversarial samples. We expect our model to achieve similar performance as the simple training approach when evaluating on real samples only. 

\subsection{Prediction Confidence}
One important thing we noticed that the proposed approach is not only able to maintain the accuracy level during an adversarial attack but also able to keep the prediction confident.
Figure~\ref{figs:boxplot} shows the prediction confidence of the proposed approach and the two baseline approaches on the real and adversarial samples. The figure reveals that on the real samples, the simple training approach generates the most confident prediction results with the predicted probability between $0.99$ to $1.0$. The proposed approach produces the second most confident prediction results (the predicted probability between $0.97$ to $1.0$). The adversarial training approach has the worst prediction confidence, which ranges between $0.9$ to $1.0$. However, when testing on the adversarial samples, the prediction confidence of the simple training decreased dramatically. The lowest prediction confidence dropped to $0.62$, which original was $0.99$ with the real samples. Similarly, the prediction confidence of the adversarial training is also decreased significantly. Unlike the baseline models, the proposed approach is able to maintain the prediction confidence at the similar level.

\begin{figure}[!tb]
  \centering
        \includegraphics[width=0.4875\textwidth]{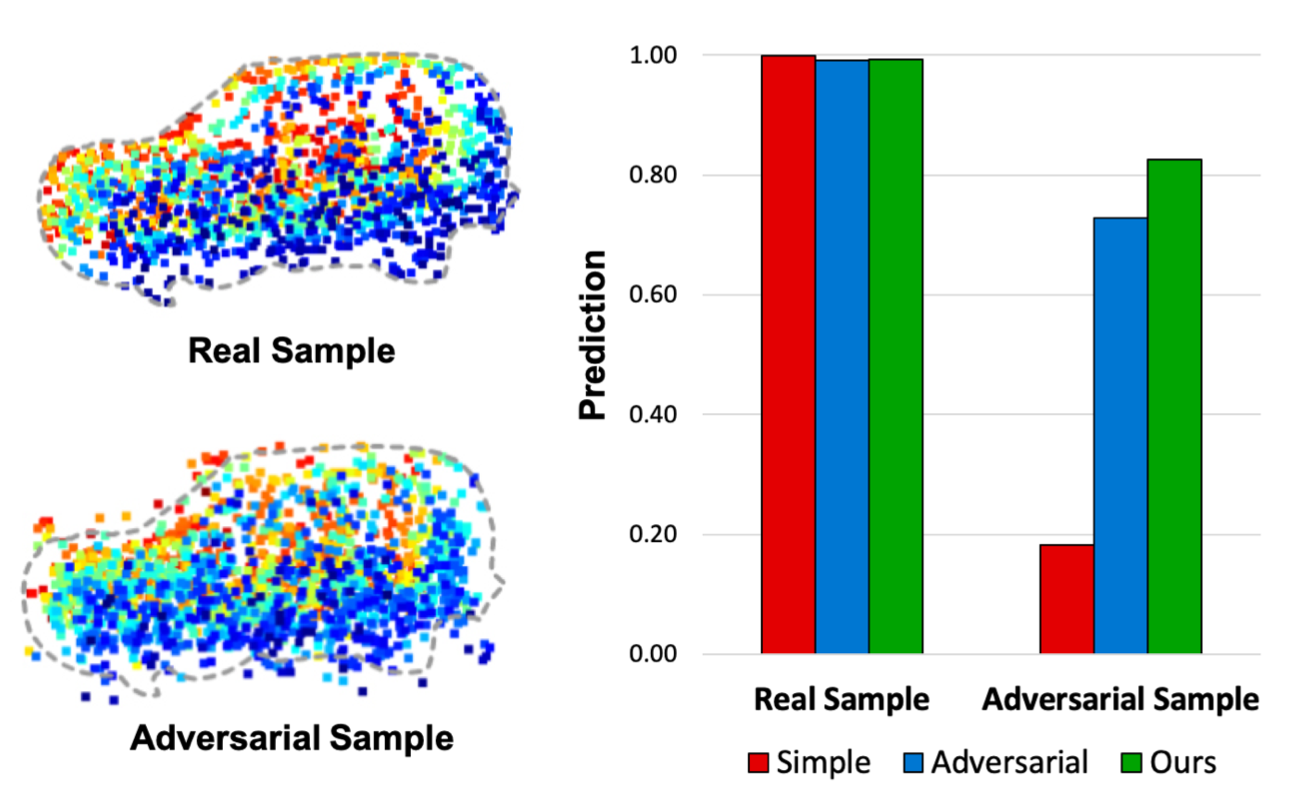}
        \caption{Classification result of \textbf{real car} sample (left-top) and \textbf{adversarial car} sample (left-bottom) of different methods.}
        \label{fig:vis_1}
\end{figure}

\begin{figure}[!tb]
  \centering
        \includegraphics[width=0.4875\textwidth]{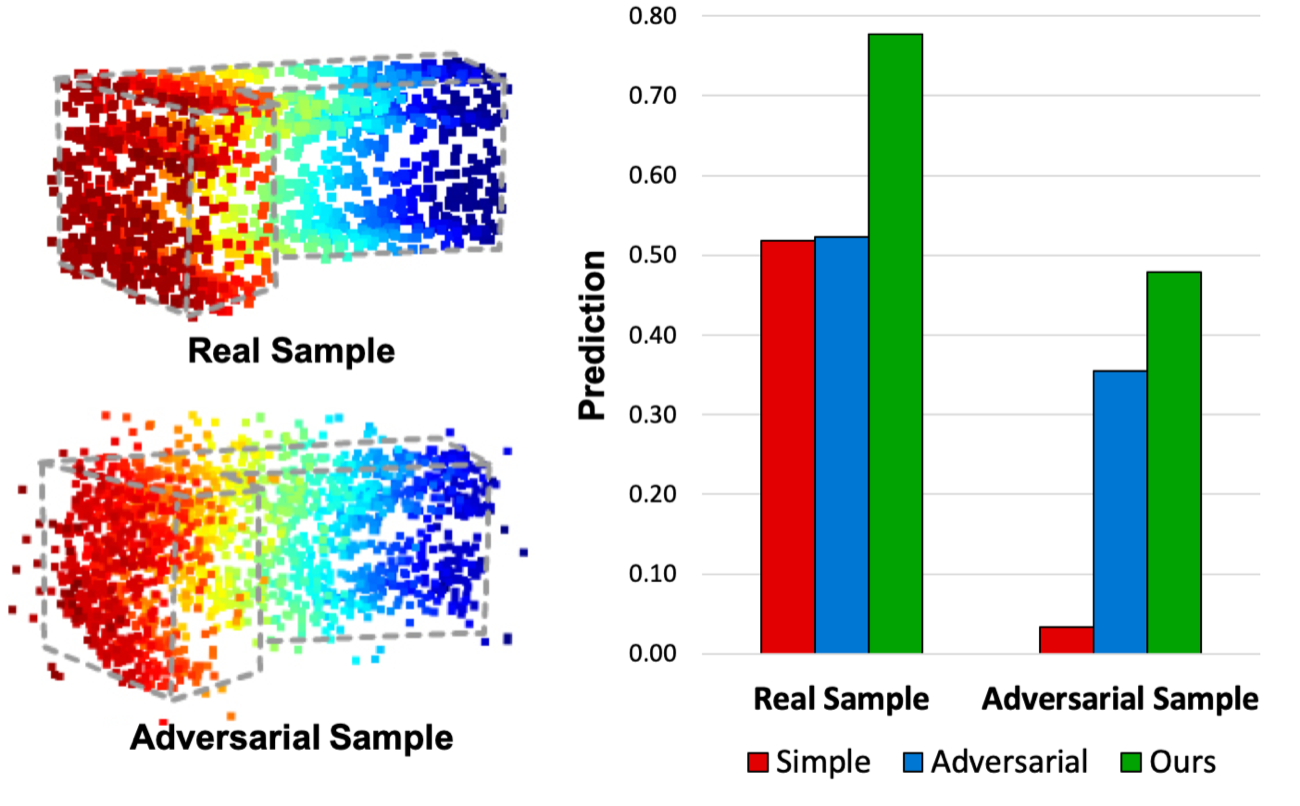}
        \caption{Classification result of \textbf{real table} sample (left-top) and \textbf{adversarial table} sample (left-bottom) of different methods.}
        \label{fig:vis_2}

    \label{fig:vis}
\end{figure}

We give two examples to show the prediction confidence. Figure~\ref{fig:vis_1} reveals that when testing with the real car sample, all the three models perform an almost perfect prediction. However, when testing on the adversarial car sample, only the proposed approach is able to maintain the prediction confidence. Figure~\ref{fig:vis_2} indicates a similar result of the real table sample and the adversarial sample.



\subsection{Feature Space Visualizations}

\begin{figure*}
  \centering
  \includegraphics[width=0.9\textwidth]{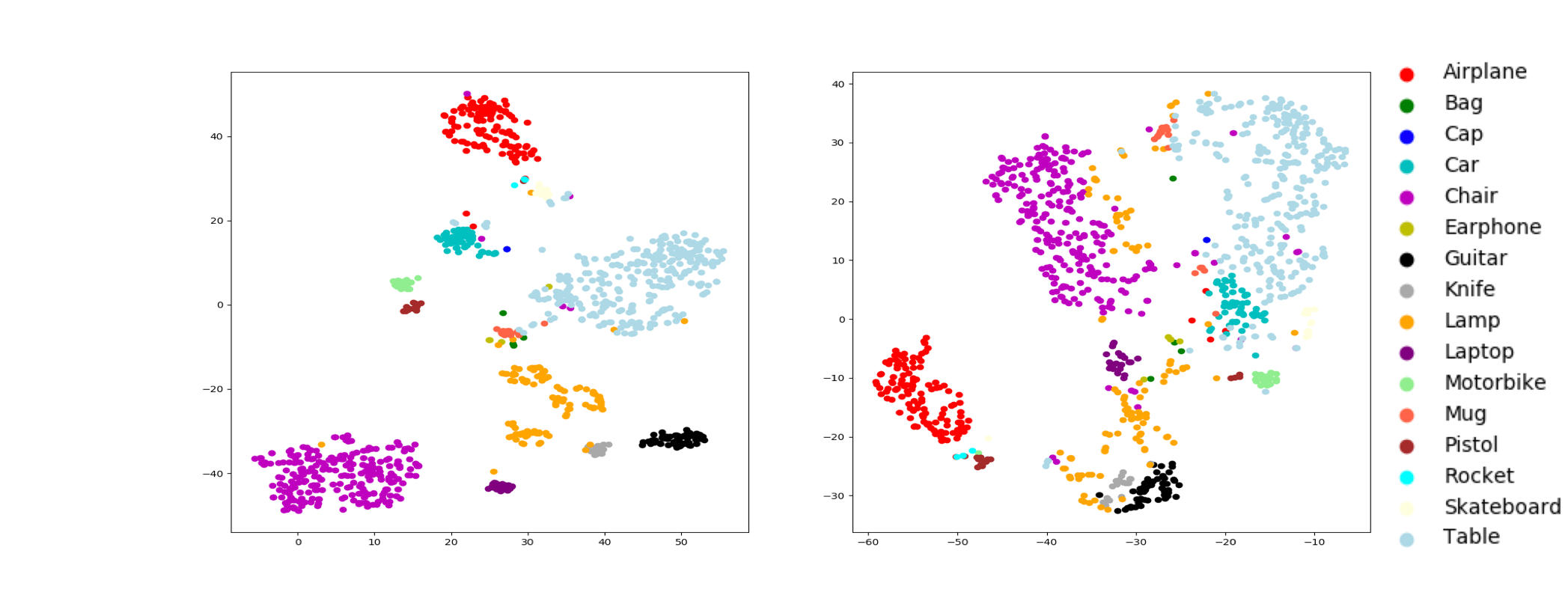}
  \caption{Two t-SNE plots, which visualize the feature cluster compactness of the simple training baseline approach. The real sample features (left) are more separate while the adversarial samples (right) have more inter-class overlap.}
  \label{figs:simple_features}
\end{figure*}

\begin{figure*}
  \centering
  \includegraphics[width=0.9\textwidth]{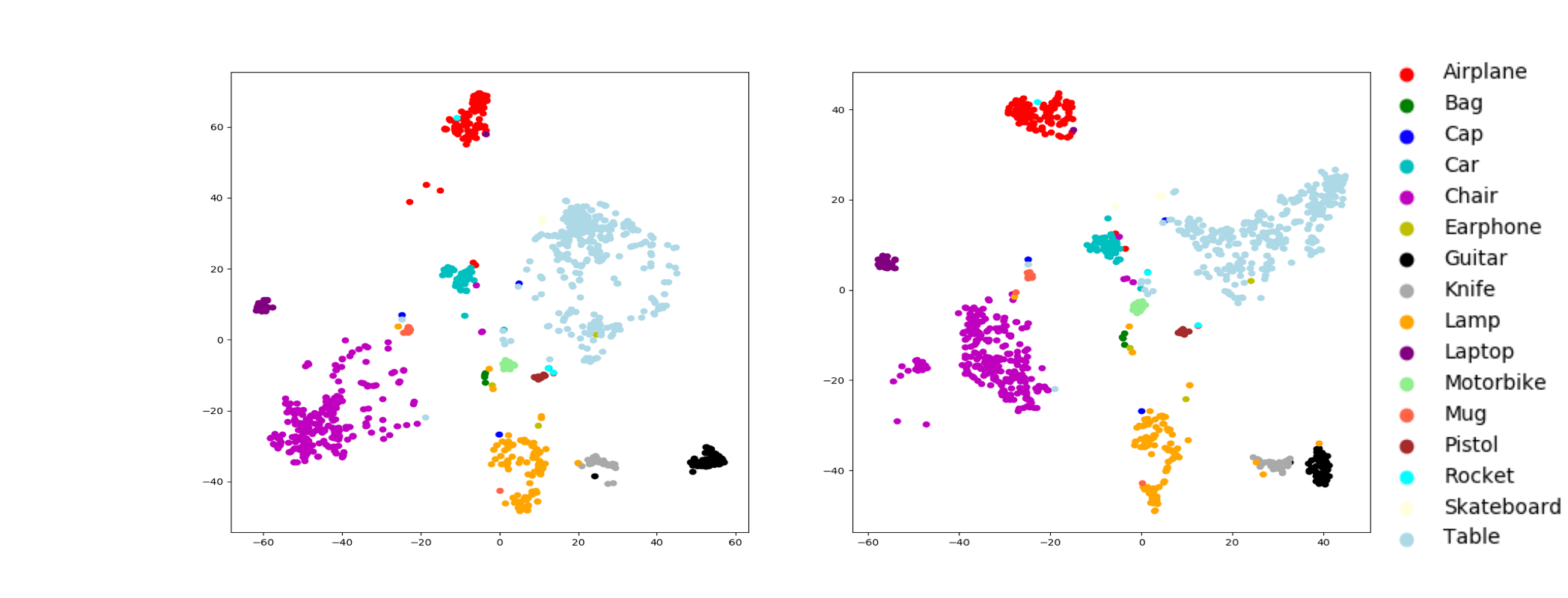}
  \caption{Two t-SNE plots, which visualize the feature cluster compactness of the Defense-PointNet approach. Our approach enhances  intra-class compactness which makes adversarial sample features (right) as separate as real sample features (left).}
  \label{figs:ours_features}
\end{figure*}

We visualize the 2-D t-SNE plots of $1024D$ latent vectors extracted by simple training model in Figure~\ref{figs:simple_features} and Defense-PointNet model in Figure~\ref{figs:ours_features}. We can see that, for simple training model, feature clusters of adversarial samples (Figure~\ref{figs:simple_features} right) are less separate and the overlap of different classes causes the success of adversarial attack. Defense-PointNet enhances intra-class compactness (Figure~\ref{figs:ours_features} right), thereby reducing the feature cluster overlap, leading to a lower adversary success for a bounded perturbation $\epsilon \leq 0.1$.


\section{Conclusion}
We introduced a novel approach for protecting PointNet against adversarial attacks. In several critical experiments, we demonstrated our proposed approach can significantly improve the robustness against adversarial samples of PointNet, as well as maintain the high prediction confidence. We also provided a interpretation of how our model affects the feature space representations by visualizing latent vectors. We hope this work can be able to provide a baseline as well as a guideline for future 3D adversarial attacks, defending strategies and interpretability research.



\section*{Acknowledgements}

We gratefully acknowledge the support of NSF CAREER (IIS-1553116).

{
\small
\bibliographystyle{IEEEtran}
\bibliography{ieee}
}

\end{document}